\documentclass{article}
\usepackage{ijcai26}

\usepackage{times}
\usepackage{soul}
\usepackage{url}
\usepackage[hidelinks]{hyperref}
\usepackage[utf8]{inputenc}
\usepackage[small]{caption}
\usepackage{graphicx}
\usepackage{amsmath}
\usepackage{amsthm}
\usepackage{amsfonts}
\usepackage{booktabs}
\usepackage{algorithm}
\usepackage{algorithmic}
\usepackage[switch]{lineno}

\usepackage[T1]{fontenc}
\usepackage[T5]{fontenc}

\usepackage[tone]{tipa}

\usepackage{multirow}
\usepackage[normalem]{ulem}
\useunder{\uline}{\ul}{}

\title{ViSpeechFormer: A Phonemic Approach for Vietnamese Automatic Speech Recognition}

\author{
Khoa Anh Nguyen$^{1,2,3}$
\and
Long Minh Hoang$^{1,2,3}$\and
Nghia Hieu Nguyen$^{1,2,3}$\and
Luan Thanh Nguyen$^{1,2,3}$\And
Ngan Luu-Thuy Nguyen$^{1,2,3}$\\
\affiliations
$^1$Faculty of Information Science and Engineering\\
$^2$University of Information Technology\\
$^3$Vietnam National University, Ho Chi Minh city, Vietnam\\
\emails
\{22520675, 22520809\}@gm.uit.edu.vn,
\{nghiangh, luannt, ngannlt\}@uit.edu.vn
}

\begin{document}

\maketitle

\begin{abstract}
Vietnamese has a phonetic orthography, where each grapheme corresponds to at most one phoneme and vice versa. Exploiting this high grapheme–phoneme transparency, we propose ViSpeechFormer (\textbf{Vi}etnamese \textbf{Speech} Trans\textbf{Former}), a phoneme-based approach for Vietnamese Automatic Speech Recognition (ASR). To the best of our knowledge, this is the first Vietnamese ASR framework that explicitly models phonemic representations. Experiments on two publicly available Vietnamese ASR datasets show that ViSpeechFormer achieves strong performance, generalizes better to out-of-vocabulary words, and is less affected by training bias. This phoneme-based paradigm is also promising for other languages with phonetic orthographies. The code will be released upon acceptance of this paper.
\end{abstract}

\section{Introduction}

Most studies in automatic speech recognition (ASR) focus on how to represent and model speech signals effectively. Early work \cite{ctc} formulated ASR as a sequence-to-sequence alignment problem, in which the input acoustic sequence is significantly longer than the output label sequence. To address this mismatch, Connectionist Temporal Classification (CTC) was introduced as an objective function that enables deep neural networks to learn alignments between multiple input frames and output tokens. Building upon this idea, the Transducer loss \cite{rnnt} was proposed as an effective alternative to CTC by conditioning each output token not only on the input signal but also on previously generated output tokens.

Subsequent studies proposed the encoder--decoder architecture \cite{seq2seq}, which can be trained using cross-entropy loss without explicitly modeling the alignment between input and output sequences. However, this architecture requires encoding the entire input sequence before generating any output tokens, making it more complex and computationally expensive than CTC-based or Transducer-based approaches, especially in streaming or low-latency scenarios.

In addition, most ASR studies evaluate their proposed methods at the word level \cite{lsvsc}, subword level \cite{conformer,zipformer,multi-convformer,tasa}, or character level \cite{rnnt,speech-transformer}. Word-level modeling suffers from the out-of-vocabulary (OOV) problem, while character-level modeling introduces significant alignment ambiguity between acoustic signals and characters. Subword-level modeling, although effective for many languages, is less suitable for isolating languages such as Vietnamese, where words are typically monosyllabic and exhibit no morphological variation \cite{hao1998,giap2008,giap2011}.

Motivated by these limitations, we propose to generate transcripts at the phoneme level for Vietnamese ASR, leveraging the high correspondence between graphemes and phonemes in the Vietnamese language. Accordingly, we introduce a Phonemic Decoder that is specifically designed to produce phoneme-level transcriptions, along with a tokenization algorithm that converts Vietnamese text into phoneme sequences and vice versa. Experimental results across various ASR architectures demonstrate that the proposed phonemic decoder is more effective for Vietnamese than character-level and word-level decoding approaches.

\section{Related Works}

Early ASR systems were primarily based on hybrid HMM--GMM or DNN--HMM architectures, where phoneme-level modeling was realized through context-dependent states and pronunciation lexicons \cite{hmm,hmm-gmm}. While these systems explicitly incorporated phonetic knowledge, they required complex pipelines and handcrafted lexicons, limiting scalability for low-resource languages such as Vietnamese. Prior Vietnamese ASR work under this paradigm emphasized tone-aware phoneme inventories, but performance was constrained by limited annotated data and tonal variability.

With the advent of end-to-end ASR, recent approaches have shifted toward character-, subword-, or word-level decoding, eliminating the need for explicit phoneme modeling. CTC-based methods \cite{ctc,rnn-asr}, attention-based encoder--decoder models \cite{las}, and hybrid CTC/attention frameworks \cite{ctc-attention}, together with Transformer and Conformer architectures \cite{conformer}, have achieved strong results across many languages, including Vietnamese. However, for tonal and low-resource languages, character- or word-level decoding can suffer from data sparsity and pronunciation ambiguity, as acoustic--phonetic regularities are learned only implicitly.

Recent studies have explored phoneme-based or phoneme-aware decoding in end-to-end frameworks, showing improved robustness and data efficiency \cite{advanced-ctc,rnn-ctc}. Self-supervised models such as wav2vec~2.0 further suggest that learned speech representations align well with phonetic units \cite{wav2vec2}. Despite these advances, explicit phoneme-level decoding for Vietnamese ASR remains underexplored. In contrast, our work introduces a phoneme-level decoder tailored to Vietnamese, enabling explicit modeling of tonal phonemes within an end-to-end framework.

\section{Methodology}

\subsection{Backgrounds} \label{sec:background}

\begin{figure}
    \centering
    \includegraphics[width=\linewidth]{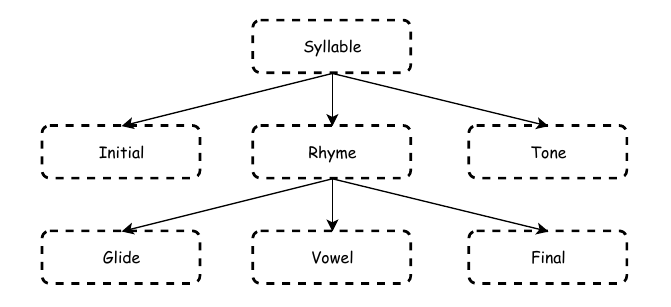}
    \caption{The structure of syllable in Vietnamese.}
    \label{fig:syllable-structure}
\end{figure}

\begin{figure}
    \centering
    \includegraphics[width=\linewidth]{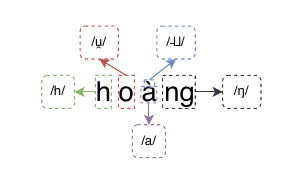}
    \caption{Example of the transparent between graphemes and phonemes in Vietnamese. The given word in this example means \textit{royal} in English.}
    \label{fig:syllable-example}
\end{figure}

Vietnamese exhibits as an isolating monosyllabic language \cite{hao1998,giap2008,giap2011} whose word and morpheme identify the same unit. To this end, every Vietnamese word has at most one syllable with stable structure: initial, rhyme, and tone \cite{hao1998,thuat2016}. Rhyme is the compulsory component and can be further analysed into glide, vowel, and final (Figure \ref{fig:syllable-structure}). Moreover, this language has a phonetic orthography where grapheme and phoneme are highly correspondent. Particularly, every phoneme in Vietnamese has consistent writing form regardless to tenses, moods, and cases. Accordingly, Vietnamese native can determine which grapheme corresponding to the respective syllabic components (Figure \ref{fig:syllable-example}). Roughly speaking, we can know how to write a Vietnamese word given its pronunciation and vice versa.

According to former phonetic studies in Vietnamese \cite{hao1998,thuat2016}, the writing form of these phonemes is consistent regardless of grammar. Vietnamese syllable has 22 phonemes for initial, one phoneme for glide, 15 phonemes for vowel, 10 phonemes for final, and 6 phonemes to indicate tone. In particular:

\begin{itemize}
    \item 6 tones are denoted by a mark above or below the graphemes of vowels:
    \begin{itemize}
        \item Flat tone is denoted by nothing (a).
        \item Low falling tone /\textipa{\tone{22}\tone{11}}/ is denoted by a grave accent (à).
        \item Mid raising tone /\textipa{\tone{33}\tone{55}}/ is denoted by an acute accent (á).
        \item Mid falling tone /\textipa{\tone{33}\tone{11}}/ is denoted by a hook above (ả).
        \item Mid glottalized-falling tone /\textipa{\tone{33}P\tone{55}}/ is denoted by a tilde above (ã).
        \item Mid glottalized-raising tone /\textipa{\tone{33}P\tone{11}}/ is denoted by a dot below (ạ).
    \end{itemize}
    In the following texts, we provide writing forms of vowels regarding the mentioned phonemes. These examples might include the tone mark above or below the graphemes. Readers can discard these marks to see the true writing form of the vowels in Vietnamese. For instance, the dot below \textbf{iê} /\textipa{i9}/ in word \textbf{kiệm} /\textipa{kiem\tone{33}P\tone{55}}/ denotes the mid glottalized-raising tone /\textipa{\tone{33}P\tone{55}}/.
    
    \item 22 initials have 26 writing forms:
    \begin{itemize}
        \item b /\textipa{b}/. Eg: \textbf{b}a mẹ, \textbf{b}ánh kẹo, \textbf{b}uôn \textbf{b}án.
        \item t /\textipa{t}/. Eg: \textbf{t}âm \textbf{t}ư, \textbf{t}ịnh \textbf{t}iến, \textbf{t}ính cách.
        \item th /\textipa{t\textsuperscript{h}}/. Eg: \textbf{th}ách \textbf{th}ức, \textbf{th}ành \textbf{th}ạo.
        \item k, c, or q /\textipa{k}/. Eg: \textbf{c}ách mạng, \textbf{q}uan hệ, hiện \textbf{k}im.
        \item ph /\textipa{f}/. Eg: \textbf{ph}ụ huynh, \textbf{ph}ong cách, \textbf{ph}ân định.
        \item đ /\textipa{d}/. Eg: \textbf{đ}ưa \textbf{đ}ón, \textbf{đ}ậm \textbf{đ}à.
        \item gh or g /\textipa{7}/. Eg: \textbf{g}a tàu, \textbf{g}ánh hát, \textbf{g}anh \textbf{gh}ét.
        \item gi /\textipa{z}/. Eg: \textbf{gi}ếng nước, \textbf{gi}ống loài.
        \item đ /\textipa{d}/. Eg: \textbf{đ}ứng \textbf{đ}ắn, \textbf{đ}êm tối, \textbf{đ}èn \textbf{đ}uốc.
        \item d /\textipa{j}/. Eg: \textbf{d}a \textbf{d}ẻ, tiêu \textbf{d}ùng, \textbf{d}ụng cụ.
        \item x /\textipa{s}/. Eg: sản \textbf{x}uất, \textbf{x}uất \textbf{x}ứ, \textbf{x}e cộ.
        \item s /\textipa{\:s}/. Eg: xác \textbf{s}uất, \textbf{s}o \textbf{s}ánh, \textbf{s}ao chép.
        \item ch /\textipa{\t{cC}}/. Eg: \textbf{ch}ứa \textbf{ch}an, \textbf{ch}e \textbf{ch}ở/.
        \item tr /\textipa{\t{t\:s}}/. Eg: \textbf{tr}anh chấp, tiệt \textbf{tr}ùng.
        \item ng or ngh /\textipa{ng}/. Eg: mong \textbf{ng}óng, tình \textbf{ngh}ĩa.
        \item nh /\textipa{\textltailn}/. Eg: \textbf{nh}à cửa, nỗi \textbf{nh}ớ, \textbf{nh}ung lụa.
        \item l /\textipa{l}/. Eg: \textbf{l}ấm \textbf{l}em, \textbf{l}ung \textbf{l}inh, \textbf{l}ối về.
        \item r /\textipa{r}/. Eg: \textbf{r}ậm \textbf{r}ạp, \textbf{r}ón \textbf{r}én, \textbf{r}ực \textbf{r}ỡ.
        \item kh /\textipa{x}/. Eg: \textbf{kh}ó \textbf{kh}ăn, \textbf{kh}ởi sắc, \textbf{kh}ấm \textbf{kh}á.
        \item v /\textipa{v}/. Eg: \textbf{v}ui \textbf{v}ẻ, \textbf{v}ương \textbf{v}ấn, \textbf{v}ẫy \textbf{v}ùng.
        \item m /\textipa{m}/. Eg: \textbf{m}ong \textbf{m}ỏi, \textbf{m}ay \textbf{m}ắn, \textbf{m}ênh \textbf{m}ông.
        \item n /\textipa{n}/. Eg: đất \textbf{n}ước, \textbf{n}úi \textbf{n}on, \textbf{n}ông cạn.
    \end{itemize}

    \item The glide /\textipa{\textsubarch{u}}/ has two writing forms: u or o. Eg: q\textbf{u}ê nhà, h\textbf{o}a cỏ, kh\textbf{u}yến khích.

    \item 03 diphthongs have 8 writing forms:
    \begin{itemize}
        \item iê, yê, ia, or ya /\textipa{ie}/. Eg: kh\textbf{iế}m thị, \textbf{yê}n ắng, ch\textbf{ia} sẻ, khu\textbf{ya} khoắt.
        \item uô or ua /\textipa{uo}/. Eg: kh\textbf{uô}n khổ, m\textbf{ua} bán.
        \item ươ or ưa /\textipa{W@}/. Eg: kh\textbf{ướ}u giác, dây d\textbf{ưa}.
    \end{itemize}

    \item 12 monophthongs have 13 writing forms:
    \begin{itemize}
        \item a /\textipa{a}/. Eg: b\textbf{a} mẹ, tr\textbf{a}nh vẽ, ng\textbf{ã} b\textbf{a}, m\textbf{ải} miết, làng ch\textbf{ài}.
        \item ă or a /\textipa{ă}/. Eg: ánh \textbf{ắ}ng, n\textbf{ắ}m tay, n\textbf{ă}m tháng, \textbf{áy} n\textbf{áy}, chạy nhảy.
        \item â /\textipa{\u{@}}/. Eg: n\textbf{â}ng niu, \textbf{ấ}n định, ng\textbf{â}n vang.
        \item i or y /\textipa{i}/. Eg: th\textbf{i} cử, tr\textbf{ĩ}u nặng, b\textbf{ĩ}u môi.
        \item ê /\textbf{e}/. Eg: k\textbf{ế}t quả, thể hi\textbf{ệ}n, mân m\textbf{ê}.
        \item e /\textipa{E}/. Eg: mùa h\textbf{è}, x\textbf{e} cộ, t\textbf{é} ngã.
        \item u /\textipa{u}/. Eg: th\textbf{u} mua, m\textbf{ủ}m mỉm, l\textbf{u}ng lay, tr\textbf{u}ng thành.
        \item ư /\textipa{W}/. Eg: tr\textbf{ư}ng cầu, xây d\textbf{ự}ng, \textbf{ư}ng ý.
        \item o /\textipa{O}/. Eg: chăm s\textbf{ó}c, m\textbf{o}ng ng\textbf{ó}ng, tr\textbf{o}ng veo.
        \item oo /\textipa{O:}/. Eg: x\textbf{oo}ng chảo.
        \item ô /\textipa{o}/. Eg: tr\textbf{ố}ng đ\textbf{ồ}ng, \textbf{ố}ng hút, c\textbf{ố} nhân.
        \item ơ /\textipa{@}/. Eg: m\textbf{ơ} mộng, c\textbf{ơ} nh\textbf{ỡ}, ch\textbf{ơ}i b\textbf{ờ}i.
    \end{itemize}

    \item 10 final consonants have 12 writing forms:
    \begin{itemize}
        \item i or y /\textipa{\textsubarch{i}}/. Eg: làng chà\textbf{i}, mỏ\textbf{i} mệt, chạ\textbf{y} đua, bay nhả\textbf{y}.
        \item m /\textipa{m}/. Eg: ê\textbf{m} ấ\textbf{m}, nhiệ\textbf{m} màu, mâ\textbf{m} cỗ.
        \item n /\textipa{n}/. Eg: na\textbf{n} giải, no\textbf{n} nớt, tả\textbf{n} mạ\textbf{n}.
        \item ng /\textipa{\ng}/. Eg: sa\textbf{ng} trọ\textbf{ng}, trố\textbf{ng} trải, su\textbf{ng} túc.
        \item nh /\textipa{\textltailn}/. Eg: nha\textbf{nh} nhẹn, bê\textbf{nh} vực, bi\textbf{nh} quyền.
        \item p /\textipa{p}/. Eg: phậ\textbf{p} phồng, thấ\textbf{p} thỏm, thá\textbf{p} tùng, gượng é\textbf{p}, ức hiế\textbf{p}, tẩm ướ\textbf{p}.
        \item t /\textipa{t}/. Eg: lấn á\textbf{t}, bá\textbf{t} đĩa, kế\textbf{t} quả, hớ\textbf{t} hả.
        \item c /\textbf{k}/. Eg: cú\textbf{c} áo, chự\textbf{c} chờ, bố\textbf{c} vá\textbf{c}.
        \item ch /\textbf{c}/. Eg: cá\textbf{ch} thứ\textbf{c}, chí\textbf{ch} ngừa.
        \item u or o /\textipa{\textsubarch{u}}/. Eg: tra\textbf{u} chuốt, chị\textbf{u} đựng, xoong chả\textbf{o}, xiêu vẹ\textbf{o}.
    \end{itemize}
\end{itemize}

However, although the conversion from grapheme to phonemes is straightforward (that is, many graphemes correspond to unique phonemes), the inversion is not all ways many-to-one mapping. There are some phonemes having one-to-many mapping with graphemes, such as the initial /\textipa{7}/ can be written as \textbf{g} or \textbf{gh}, or the diphthong /\textipa{ie}/ can be written as \textbf{iê, yê, ia} or \textbf{ya}. Actually, a particular writing form of such phonemes is determined consistently via the neighbor phonemes or graphemes following the orthographic rules.  In particular:s

\begin{itemize}
    \item the diphthong /\textipa{ie}/ is written as:
    \begin{itemize}
        \item \textbf{iê} if the rhyme has a final consonant and no glide. Eg: k\textbf{iến} thức, tiết k\textbf{iệm}.
        \item \textbf{yê} if the rhyme has a final consonant and the glide written as \textbf{u}. Eg: kh\textbf{uyên} bảo, \textbf{uyển} ch\textbf{uyển}, câu ch\textbf{uyện}.
        \item \textbf{ya} if the rhyme has no final consonant and the glide written as \textbf{u}. Eg: đêm kh\textbf{uya}.
        \item \textbf{ia} if the rhyme has no final consonant and no glide. Eg: b\textbf{ìa} sách, ch\textbf{ia} sẻ.
    \end{itemize}

    \item The diphthong /\textipa{uo}/ is written as:
    \begin{itemize}
        \item \textbf{uô} if the rhyme has a final consonant. Eg: nỗi b\textbf{uồn}, m\textbf{uối} biển, m\textbf{uộn} màn, ch\textbf{uồn} ch\textbf{uồn}.
        \item \textbf{ua} if the rhyme has no final consonant. Eg: ch\textbf{ùa} chiền, nhảy m\textbf{úa}.
    \end{itemize}

    \item The initial /\textipa{k}/ is written as:
    \begin{itemize}
        \item \textbf{k} if followed by \textbf{i} /\textipa{i}/, \textbf{ê} /\textipa{e}/, \textbf{e} /\textipa{E}/, \textbf{iê} /\textipa{ie}/. Eg: \textbf{kị}p thời, \textbf{kiể}u cách, \textbf{kè}m cặp.
        \item \textbf{q} if followed by \textbf{u} /\textipa{\textsubarch{u}}/ as the glide. Eg: \textbf{qu}ê hương, \textbf{qu}à cáp.
        \item \textbf{c} otherwise. Eg: \textbf{cứ}ng \textbf{cỏ}i, \textbf{cở}i mở, hạt \textbf{cá}t, rau \textbf{củ}.
    \end{itemize}

    \item The initial /\textipa{7}/ is written as:
    \begin{itemize}
        \item \textbf{gh} if followed by \textbf{i} /\textipa{i}/, \textbf{ê} /\textipa{e}/, \textbf{e} /\textipa{E}/, \textbf{iê} /\textipa{ie}/. Eg: bàn \textbf{ghế}, \textbf{ghi} chép, \textbf{ghe} tàu.
        \item \textbf{g} otherwise. Eg: thanh gươm, gồng gánh, gọi điện.
    \end{itemize}

    \item The initial /\textipa{\ng}/ is written as:
    \begin{itemize}
        \item \textbf{ngh} if followed by textbf{i} /\textipa{i}/, \textbf{ê} /\textipa{e}/, \textbf{e} /\textipa{E}/, \textbf{iê} /\textipa{ie}/. Eg: \textbf{nghe} ngóng, \textbf{nghiê}m nghị, \textbf{nghệ} sĩ.
        \item \textbf{ng} otherwise. Eg: \textbf{ngà}nh nghề, \textbf{ngỗ} nghịch, \textbf{ngọ}t \textbf{ngà}o.
    \end{itemize}

    \item The diphthong /\textipa{ie}/ is written as:
    \begin{itemize}
        \item \textbf{iê} if the rhyme has a final consonant and no glide. Eg: k\textbf{iến} thức, tiết k\textbf{iệm}.
        \item \textbf{yê} if the rhyme has a final consonant and the glide written as \textbf{u}. Eg: kh\textbf{uyên} bảo, \textbf{uyển} ch\textbf{uyển}, câu ch\textbf{uyện}.
        \item \textbf{ya} if the rhyme has no final consonant and the glide written as \textbf{u}. Eg: đêm kh\textbf{uya}.
        \item \textbf{ia} if the rhyme has no final consonant and no glide. Eg: b\textbf{ìa} sách, ch\textbf{ia} sẻ.
    \end{itemize}

    \item The diphthong /\textipa{uo}/ is written as:
    \begin{itemize}
        \item \textbf{uô} if the rhyme has a final consonant. Eg: nỗi b\textbf{uồn}, m\textbf{uối} biển, m\textbf{uộn} màn, ch\textbf{uồn} ch\textbf{uồn}.
        \item \textbf{ua} if the rhyme has no final consonant. Eg: ch\textbf{ùa} chiền, nhảy m\textbf{úa}.
    \end{itemize}

    \item The diphthong /\textipa{W@}/ is written as:
    \begin{itemize}
        \item \textbf{ươ} if the rhyme has a final consonant. Eg: bia r\textbf{ượu}, h\textbf{ưởng} thụ, chiêm ng\textbf{ưỡng}.
        \item \textbf{ua} if the rhyme has no final consonant. Eg: ch\textbf{ùa} chiền, nhảy m\textbf{úa}.
    \end{itemize}

    \item The monophthong /\textipa{ă}/ is written as:
    \begin{itemize}
        \item \textbf{a} if is is followed by character \textbf{y}. Eg: m\textbf{áy} bay, c\textbf{ay} nồng, t\textbf{ay} chân.
        \item \textbf{ă} otherwise. Eg: b\textbf{ắt} tay, b\textbf{ằng} lòng, may m\textbf{ắn}.
    \end{itemize}

    \item The monophthong /\textipa{i}/ is written as:
    \begin{itemize}
        \item \textbf{i} if the rhyme has a final consonant. Eg: l\textbf{íu} r\textbf{ít}.
        \item \textbf{y} if the rhyme has no consonant. Eg: k\textbf{ỷ} luật, l\textbf{ý} do.
    \end{itemize}

    \item The final /\textipa{\textsubarch{i}}/ is written as:
    \begin{itemize}
        \item \textbf{i} if the rhyme has the front vowel /\textipa{a}/, the central vowel /\textipa{W, W9}/ or the back vowels /\textipa{u, o, O, uo}/ as the vowel. Eg: m\textbf{ải} mê, m\textbf{ui} thuyền, h\textbf{ỏi} han, m\textbf{ồi} chài, g\textbf{ửi} gắm, l\textbf{ười} biếng, n\textbf{uôi} nấng.
        \item \textbf{y} if the rhyme has /\textipa{\u{a}, \u{9}}/ as the vowel. Eg: b\textbf{ay} lượn, ch\textbf{ạy} nh\textbf{ảy}, c\textbf{ấy} c\textbf{ày}.
    \end{itemize}
\end{itemize}

A comprehensive details of orthography rules in Vietnamese can be found in \cite{thuat2016}. Following the these rules, there is no ambiguity in converting phonemes to graphemes, making them an one-to-one correspondence. Analysis in this Section serves as the fundamentals for our ViPhonER Tokenization algorithm, which is described comprehensively in Section \ref{sec:viphoner}.

\subsection{ViPhonER - Vietnamese Phonemic Tokenization Algorithm} \label{sec:viphoner}

Following the linguistic features of Vietnamese mentioned in Section \ref{sec:background}, we develop a tokenization algorithm, ViPhonER (\textbf{Vi}etnamese \textbf{Phon}emic tokeniz\textbf{ER}), for converting Vietnamese text into the sequence of IPA-represented phonemes. The pseudo code for our tokenization algorithm is given in Alg. \ref{alg:text2phoneme}.

\begin{algorithm}[t]
\caption{The algorithm for converting text to phonemes.} \label{alg:text2phoneme}

\textbf{Input:} Transcript of the audio $w = (w_1, w_2, ..., w_n)$.\\

\textbf{Output:} A sequence of syllables $p = (p_1, p_2, ..., p_n)$ of the given input transcript $w = (w_1, w_2, ..., w_n)$. Each phoneme $p_i = (p_i^{init}, p_i^{glide}, p_i^{vowel}, p_i^{final}, p_i^{tone})$ is a triplet of IPA for the initial, rhyme, and tone.\\

\begin{algorithmic}[1] 

    \STATE phonemes $\leftarrow$ an empty list [];
    \STATE Let $t=0$.
    \FOR{condition}
    \STATE $p_{tone}, W$ $\leftarrow$ \text{get-tone}($W$);
    
    \STATE $p_{initial}, W$ $\leftarrow$ \text{get-initial}($W$);
    
    \STATE $p_{glide}, W$ $\leftarrow$ \text{get-glide}($W$);
    
    \STATE $p_{vowel}, W$ $\leftarrow$ \text{get-vowel}($W$);
    
    \STATE $p_{final}$ $\leftarrow$ \text{get-final}($W$);

    \STATE $p_{rhyme} \leftarrow p_{glide} \oplus p_{vowel} \oplus p_{final}$;

    \STATE phonemes $\leftarrow$ Append $(p_{init}, p_{rhyme}, p_{tone})$;
    \ENDFOR
    \STATE \textbf{return} phonemes;
\end{algorithmic}

\end{algorithm}

In particular, the ViPhonER tokenizer represents each Vietnamese word as a vector of three syllabic components: initials, rhymes, and tones. Therefore its vocabulary has 22 tokens for initials, 145 tokens for rhymes and 6 tokens for tones, which results in \textbf{the vocabulary size of 163 token}. With this small and compact size vocabulary, the decoder of our proposed method has fewer parameters compared to other baselines. Detailed statistics is given in Table \ref{tab:results} in Section \ref{sec:results}.

\begin{algorithm}[t]
\caption{The algorithm for determining the phonemes of a given Vietnamese word.} \label{alg:phoneme}

\textbf{Input:} A Vietnamese word $W$.\\

\textbf{Output:} An IPA representation of the expected phoneme (initial, vowel, tone) of the given word $W$.\\

\begin{algorithmic}[1] 

    \STATE $G \leftarrow$ set of all graphemes;
    \FOR{$g$ in $G$}

    \IF{$W$ starts with $g$}
        \STATE $p \leftarrow$ the IPA symbol of $g$;
        \STATE remove $g$ from the start of $W$;
        \STATE \textbf{break};
    \ENDIF
    
    \ENDFOR

    $p \leftarrow$ None;
    
    \STATE \textbf{return} $p$;
\end{algorithmic}

\end{algorithm}

The implementation of each method \textit{get-tone}, \textit{get-initial}, \textit{get-glide}, \textit{get-vowel}, \textit{get-final} shares the same mechanism indicated in Alg. \ref{alg:phoneme}. To this end, the ViPhonER has linear-time complexity $\mathcal{O}(n)$ where $n$ is the number of graphemes corresponding to components of Vietnamese syllable. In practice, as we analyzed in Section \ref{sec:background}, Vietnamese has 26 graphemes for initials, 01 graphemes for glides, 15 graphemes for vowels, 10 graphemes for final, and 6 marks for tones, hence the maximum value that $n$ can take is $58$. The conversion from IPA symbols to texts is straightforward thanks to the high transparency between graphemes and phonemes in Vietnamese.

\subsection{ViSpeechFormer - Vietnamese Speech Transformer}

\begin{figure*}[t]
    \centering
    \includegraphics[width=0.75\linewidth]{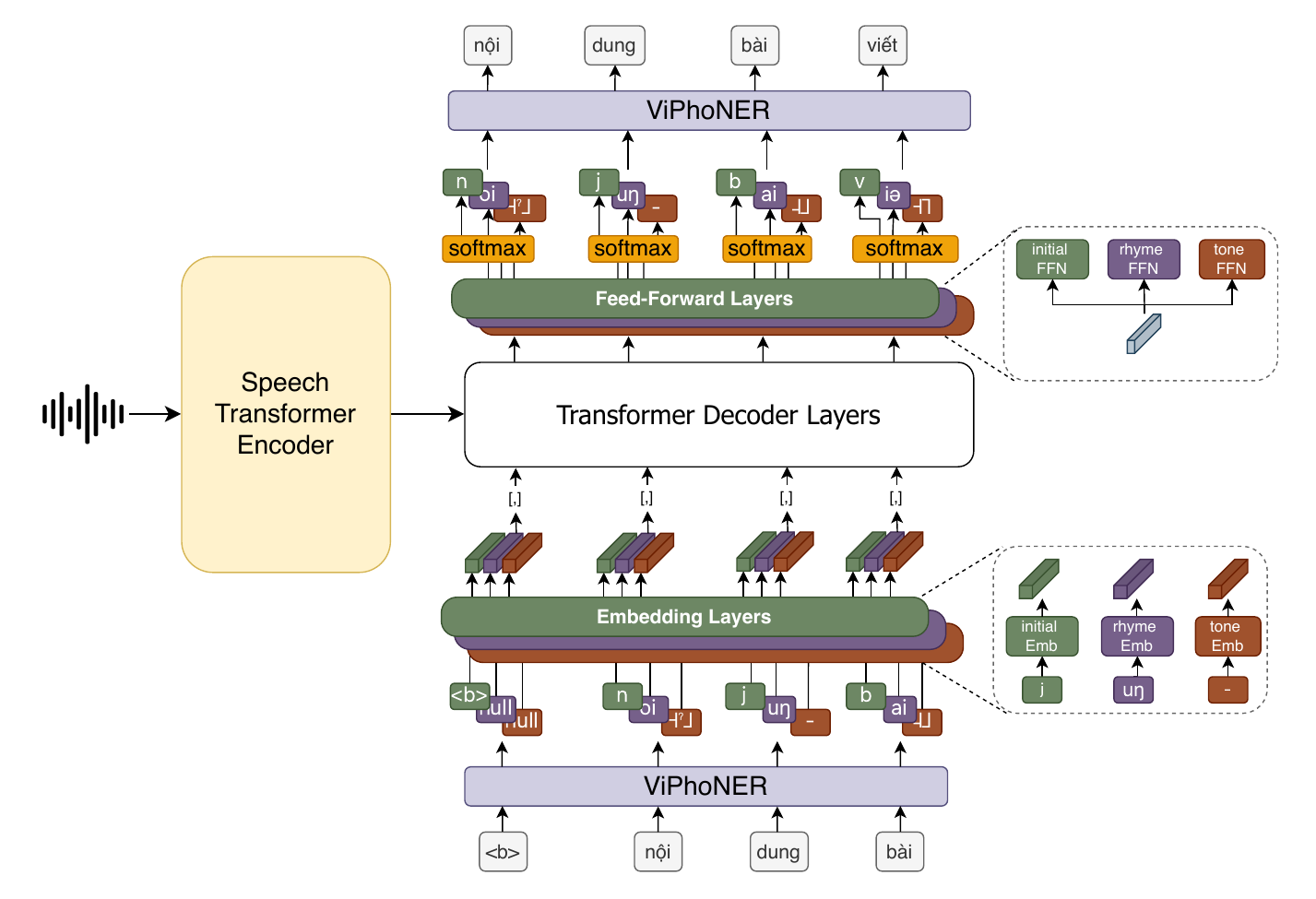}
    \caption{The architecture of the proposed ViSpeechFormer method. \textbf{FC} stands for \textbf{Fully Connected Layer}. \textbf{FFN} stands for \textbf{Feed-Forward Network}.}
    \label{fig:architecture}
\end{figure*}

Motivated by the high correspondence between graphemes and phonemes and the stable structure of syllable in Vietnamese, we propose ViSpeechFormer to model Vietnamese speech at phonemic level rather than at word or character level. In particular, our method has two componennts: Speech Transformer Encoder which is the encoder module of Speech Transformer \cite{speech-transformer} and Phonemic Decoder (Figure \ref{fig:architecture}).

The Phonemic Decoder (Figure \ref{fig:architecture}) is built on the Decoder of vanilla Transformer \cite{attention}. This module includes the Transformer Decoder layers followed by three Feed-Forward networks. Let $f_{t}^{dec}$ is the feature vectors representing the syllable at step $t$, we design three Feed-Forward networks (FFN), each of them determine the corresponding feature vectors of initials, vowels, and tones for the syllable at step $t$:

\begin{equation}
    f^{init}_t = FFN(f_{t}^{dec})
\end{equation}

\begin{equation}
    f^{vowel}_t = FFN(f_{t}^{dec})
\end{equation}

\begin{equation}
    f^{tone}_t = FFN(f_{t}^{dec})
\end{equation}

\begin{figure}[t]
    \centering
    \includegraphics[width=0.5\linewidth]{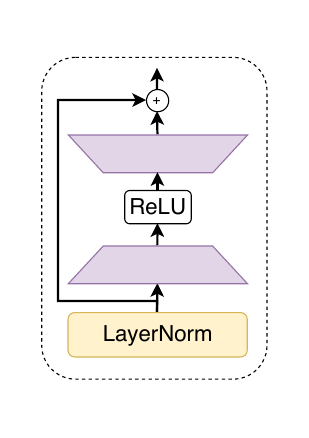}
    \caption{The architecture of Feed-Forward Network.}
    \label{fig:FFN}
\end{figure}

The Feed-Forward Network consists of the a layer normalization \cite{layernorm} followed be two fully-connected layers for projecting the feature vectors to higher dimension and scaling them back to the dimension of the input features:

\begin{equation}
    f^{p}_t = LayerNorm(f^p_t) \in \mathbb{R}^d
\end{equation}
with $p \in \{init, vowel, tone\}$, and
\begin{equation}
    f^p_t = f^p_t + W_d^T ReLU(W_u^T f_t^p) \in \mathbb{R}^d
\end{equation}
where $W_u \in \mathbb{R}^{d \times 2d}$ is the learnable parameters representing the linear function for projecting the feature vector $f_t^p$ to higher dimension and $W_d \in \mathbb{R}^{2d \times d}$ represents the linear projection to map the feature $f_t^p$ to the space of $d-$dimension (Figure \ref{fig:FFN}).

Is it crucial to note that although we tokenize each word as a tuple of three tokens, the length of the output transcript does not change because our Phonemic Decoder generates a whole vector of three dimension rather than a token. After having 3-dimension vectors of syllabic components, ViPhonER detokenizes these vector back to Vietnamese words.

In the input flow of the Phonemic Decoder, we deploy the same mechanism. Denoted the predicted 3-dimension vector from previous step $t-1$ as $p_{t-1} = [p^{init}_{t-1}, p^{rhyme}_{t-1}, p^{tone}_{t-1}]^T$, we achieved the embedded vector for each syllabic component via the Embedding Layers as:

\begin{equation}
    e_{t-1}^{init} = EmbLayer_{init}(p^{init}_{t-1}) \in \mathbb{R}^d
\end{equation}

\begin{equation}
    e_{t-1}^{rhyme} = EmbLayer_{rhyme}(p^{rhyme}_{t-1}) \in \mathbb{R}^d
\end{equation}

\begin{equation}
    e_{t-1}^{tone} = EmbLayer_{tone}(p^{tone}_{t-1}) \in \mathbb{R}^d
\end{equation}

The final embedded vector represent for the syllable at previous step $t-1$ is determined by concatenate these three embedded vectors of syllabic components:

\begin{equation}
    f^{emb}_{t-1} = [fe_{t-1}^{init}; fe_{t-1}^{rhyme}; fe_{t-1}^{tone}] \in \mathbb{R}^{3d}
\end{equation}
where $[;]$ indicates the last dimension concatenation operator. At this stage, a linear projection is applied to map this embedded vector $f^{emb}_{t-1}$ to the space of $d-$dimension:

\begin{equation}
    f^{emb}_{t-1} = W_e^T f^{emb}_{t-1}
\end{equation}
where $W_e \in \mathbb{R}^{3d \times d}$ is the learnable parameters.

\subsection{Objective Function}

Vanilla language model has Cross Entropy (CE) loss as the objective function which is defined for each output sentence $y = \{y_1, y_2, ..., y_N\}$ as follows:

$$
    CE = -\frac{1}{N} \sum_{n=1}^N log (\hat{y}_n)
$$
where $\hat{y} = \{ \hat{y}_1, \hat{y}_2, ..., \hat{y}_n \}$ is the predicted token of the language model. 

Nevertheless, the proposed Phonemic Decoder has three heads for each step of generating token, hence the language models having this decoder module require a CE objective function for each head:

\begin{equation}
    CE_{p} = -\frac{1}{N} \sum_{n=1}^N log (\hat{y}_n^{p})
\end{equation}
where $p \in \{ init, vowel, tone \}$ indicate the class of the respective syllabic component and $y^p = \{ \hat{y}_1^p, \hat{y}_2^p, ..., \hat{y}_n^p \}$ indicates the sequence of the phonemes of the component $p$.

The final loss is determined as the sum of these three losses which is:

\begin{equation}
    L = \sum_{p \in P} CE_p = -\frac{1}{N} \sum_{p \in P} \sum_{n=1}^N log (\hat{y}_n^{p})
\end{equation}
where $P = \{ init, vowel, tone \}$ indicates the class of syllabic component.

\section{Experiments}

\subsection{Datasets, Metrics and Baselines}

To prove the effectiveness of the ASR baselines using our proposed Phonemic Decoder, we conducted experiments on two open datasets for Vietnamese ASR: ViVOS \cite{vivos} and LSVSC \cite{lsvsc}. The ViVOS dataset has 15 hours of audio with 12,420 transcripts, while the LSVSC dataset has larger size containing 100 audio hours and 56,824 transcripts. We followed the previous studies \cite{conformer,speech-transformer,tasa,lsvsc} to have CER and WER as the evaluation metrics at character and word level. 
Moreover, to evaluate the performance of the proposed Phonemic Decoder, we also deployed the Phoneme Error Rate (PER) which share the same calculation as CER and WER but at phoneme level.

We conducted experiments on recent baselines for ASR task on ViVOS and LSVSC datasets. These baselines can be grouped into two classes: 
\begin{itemize}
    \item Character-level method which is Speech Transformer \cite{speech-transformer}.
    \item Subword-level methods which are Conformer \cite{conformer}, ZipFormer \cite{zipformer}, Multi-ConvFormer \cite{multi-convformer}, Conv-Transformer \cite{conv-transformer}, and TASA \cite{tasa}.
\end{itemize}
Each baseline represents its interesting idea of modeling speech signal from the output for transcription. Our experiments will show that Phonemic Decoder can be integrated with various Encoder module to tackle Vietnamese ASR task and obtain similar or better performance compared to their original decoding approach.

\subsection{Configuration}

\begin{table}[]
    \centering
    \begin{tabular}{lcccc}
    \hline
    \textbf{Dataset} & \textbf{Train} & \textbf{Dev} & \textbf{Test} & \textbf{Overall} \\ \hline
    ViVOS & 0.87 & - & 0.79 & 0.70 \\
    LSVSC & 9.19 & 9.98 & 8.89 & 9.24 \\ \hline
    \end{tabular}
    \caption{Percentage of transcripts having non-Vietnamese words in the ViVOS and LSVSC datasets.}
    \label{tab:non-vietnamese}
\end{table}

\begin{table}[t]
    \centering
    \begin{tabular}{clcccc}
    \hline
    \textbf{\#} & \textbf{Dataset} & \textbf{PER-i} & \textbf{PER-r} & \textbf{PER-t} & \textbf{PER} \\ \hline
    1 & ViVOS & 12.18 & 20.52 & 13.39 & 15.42\\
    2 & LSVSC & 6.21 & 7.01 & 5.77 & 6.39\\ \hline
    \end{tabular}
    \caption{Phoneme Error Rate (PER) of our proposed ViSpeechFormer. \textbf{PER-i}, \textbf{PER-r}, \textbf{PER-t} denotes PER of initials, rhymes, and tones, respectively.}
    \label{tab:per}
\end{table}

All baselines were trained using Adam \cite{adam} as the optimizer with the learning rate of $5 \times 10^{-5}$. On the LSVSC dataset, we configured the Encoder module of every baseline to have 12 layers while the their decoder module has 2 layers. On the ViVOS dataset, all methods has 4 layer in encoder and 1 layer in decoder. Our experiments were performed on an NVIDIA A100 80GB GPU. The training process followed the early-stopping mechanism which is interrupted if there is no improvement after 10 consecutive epochs.

Moreover, to describe how effective our ViSpeechFormer enhances on giving OOV words, \textbf{the vocabulary of all methods in our experiments were constructed on the train set only}. In particular, we followed the same configuration from previous studies \cite{speech-transformer,conformer,conv-transformer,multi-convformer,tasa,zipformer} to evaluate Speech Transformer on character level while Conformer, ZipFormer, Conv-Transformer, Multi-ConvTransformer, and TASA on subword level. For subword tokenization method, we applied the BPE tokenizater which has the vocabulary of 5000 tokens using SentencePiece toolkit on the train set of ViVOS and LSVSC datasets. 

\subsection{Data Preparation}

We mainly concentrate on Vietnamese words only, hence the ViVOS and LSVSC were preprocessed to ignore all samples whose transcript contains non-Vietnamese words. In particular, our statistics indicate that there are $9.24\%$ of samples in LSVSC including non-Vietnamese words, while in the ViVOS this percentage is $0.70\%$. These samples were discarded in our experiments. The detailed statistics of samples having non-Vietnamese words is indicated in Table \ref{tab:non-vietnamese}.

\section{Experimental Results} \label{sec:results}

\begin{table*}[htp]
    \centering
    \begin{tabular}{clcccccc}
    \hline
    \multicolumn{1}{l}{\multirow{2}{*}{\textbf{\#}}} & \multirow{2}{*}{\textbf{Baseline}} & \multicolumn{1}{l}{\multirow{2}{*}{\textbf{Level}}} & \multicolumn{1}{l}{\multirow{2}{*}{\textbf{\# Parameter of Decoder}}} & \multicolumn{2}{c}{\textbf{ViVOS}} & \multicolumn{2}{c}{\textbf{LSVSC}} \\ \cline{5-8} 
    \multicolumn{1}{l}{} &  & \multicolumn{1}{l}{} & \multicolumn{1}{l}{} & \textbf{CER} & \textbf{WER} & \textbf{CER} & \textbf{WER} \\ \hline
    1 & Conv-Transformer & subword & 2,559,089 & 16.23 & 32.69 & 07.43 & 12.59 \\
    2 & Conformer & subword & 4,302,032 & 22.87 & 37.61 & 10.61 & 15.71 \\
    3 & ZipFormer & subword & 4,302,032 & 26.34 & 38.87 & 08.88 & 13.33 \\
    4 & Multi-ConvFormer & subword & 4,302,032 & 30.98 & 44.80 & 10.58 & 15.77 \\
    5 & TASA & subword & 2,605,041 & 21.10 & 34.70 & 06.73 & 10.62 \\
    6 & Speech Transformer & character & 1,249,920 & 18.54 & 34.83 & 06.04 & 11.16 \\ \hline
    7 & ViSpeechFormer (ours) & phoneme & 2,007,892 & \textbf{11.96} & \textbf{30.49} & \textbf{05.30} & \textbf{10.39} \\ \hline
    \end{tabular}
    \caption{Results of baselines and our ViSpeechFormer method on the ViVOS and LSVSC dataset.}
    \label{tab:results}
\end{table*}

According to Table \ref{tab:results}, ViSpeechFormer attains a CER of \textbf{11.96\%} and a WER of \textbf{30.49\%}, outperforming all character- and subword-based baselines by a clear margin on the ViVOS dataset. Notably, several baseline models employ substantially larger decoders, yet still yield higher error rates. This observation suggests that decoder capacity alone is insufficient to address the challenges of Vietnamese ASR. Instead, decoding at the phoneme level allows the model to more effectively capture acoustic--phonetic correspondences, leading to improved recognition accuracy even with fewer decoder parameters.

Consistent improvements are also observed on the LSVSC dataset, where ViSpeechFormer achieves the lowest CER (\textbf{5.30\%}) and WER (\textbf{10.39\%}) among all compared methods (Table \ref{tab:results}). Despite its relatively compact decoder, the proposed model demonstrates strong generalization across datasets, indicating improved data efficiency. We attribute these gains to the explicit modeling of Vietnamese phonemes, which reduces pronunciation ambiguity and alleviates data sparsity commonly encountered in word- or subword-level decoding.

We further provided in Table \ref{tab:per} the PER of ViSpeechFormer to show its performance at phoneme level. Overall, the experimental results validate the effectiveness of phoneme-level decoding for Vietnamese ASR and highlight the advantages of incorporating language-specific phonetic knowledge into end-to-end models.

\section{Experimental Analysis}

\subsection{Analysis on Ability of Giving Correct Words}

\begin{table}[t]
    \centering
    \renewcommand{\arraystretch}{1.5} 
    \resizebox{0.5\textwidth}{!}{
    \begin{tabular}{clccc}
    \hline
    \textbf{\#} & \textbf{Method} & \textbf{\begin{tabular}[c]{@{}c@{}}Unique \\ Correct Words \end{tabular}} & \textbf{$r$} & \textbf{$\rho$} \\ \hline
    1 & Conv-Transformer & 2,393 & 23.99 & 42.69 \\
    2 & Conformer & 1,736 & 41.00 & 87.29 \\
    3 & ZipFormer & 2,367 & 26.00 & 52.17 \\
    4 & Multi-ConvFormer & 1,717 & 39.12 & 87.79 \\
    5 & TASA & 2440 & 23.12 & 38.99 \\
    6 & Speech Transformer & \textbf{2,541} & 22.36 & 30.65 \\ \hline
    7 & ViSpeechFormer (ours) & \underline{2,498} & \textbf{21.54} & \textbf{29.54} \\ \hline
    \end{tabular}}
    \caption{The number of unique correct words and the correlation (\%) between unique correct words on the train set and those predicted by the baselines and our ViSpeechFormer on the test set of LSVSC dataset. $r$ is the Pearson correlation while $\rho$ is the Spearman correlation.}
    \label{tab:correct-word-analysis}
\end{table}

Table \ref{tab:correct-word-analysis} analyzes the ability of different models to correctly generate \emph{unique words} on the test set and examines how strongly this ability depends on the word frequency distribution in the training data, measured using Pearson and Spearman correlation coefficients. Character- and subword-based baselines generally exhibit relatively high correlation scores, indicating that their correct word generation is strongly biased toward high-frequency training words. 

In particular, Conformer and Multi-ConvFormer show very high Spearman correlations (87.29\% and 87.79\%), suggesting limited generalization to low-frequency or rare words. In contrast, ViSpeechFormer achieves a competitive number of unique correct words (2,498), comparable to the best-performing baselines, while exhibiting the \emph{lowest} Pearson (21.54\%) and Spearman (29.54\%) correlations among all methods. This indicates that our phoneme-level decoder is less dependent on the training word distribution and better able to generalize beyond frequent lexical items. These results support the hypothesis that phoneme-level decoding encourages compositional word generation based on acoustic--phonetic structure rather than memorization of frequent word units, leading to improved robustness to data sparsity and long-tail vocabulary effects in Vietnamese ASR.

\subsection{Analysis on Giving Correct Out-of-Vocab Words}

\begin{table}[htp]
    \renewcommand{\arraystretch}{1.1} 
    \begin{tabular}{clc}
    \hline
    \textbf{\#} & \textbf{Method} & \textbf{Correct OOV word} \\ \hline
    1 & Conv-Transformer & 12.53 \\
    2 & Conformer & 03.03 \\
    3 & ZipFormer & 03.03 \\
    4 & Multi-ConvFormer & 02.27 \\
    5 & TASA & 14.39 \\ 
    6 & Speech Transformer & 13.64 \\ \hline
    7 & ViSpeechFormer (ours) & \textbf{27.27} \\ \hline
    \end{tabular}
    \caption{Percentage of OOV words on the LSVSC test set generated correctly of baselines and our ViSpeechFormer method.}
    \label{tab:oov}
\end{table}

Table \ref{tab:oov} reports the percentage of out-of-vocabulary (OOV) words on the test set that are generated correctly by each method. Word- and subword-based baselines generally exhibit limited OOV recognition ability, as their decoders are constrained by the vocabulary observed during training. In particular, Conformer and Multi-ConvFormer correctly generate only 3.03\% and 2.27\% of OOV words, respectively. Although some baselines such as TASA and Speech Transformer show moderate OOV performance, their accuracy remains substantially lower than that of ViSpeechFormer. 

In contrast, our proposed phoneme-level decoder achieves a markedly higher OOV accuracy of \textbf{27.27\%}, nearly doubling the best baseline result. These results demonstrate that phoneme-level decoding enables ViSpeechFormer to construct unseen words compositionally from phonetic units, substantially improving robustness to vocabulary mismatch in Vietnamese ASR.

\subsection{Analysis on Speed of Inference}

\begin{table}[htp]
    \renewcommand{\arraystretch}{1.1} 
    \begin{tabular}{clc}
    \hline
    \textbf{\#} & \textbf{Method} & \textbf{Avg. Speed of Inference} \\ \hline
    1 & Conv-Transformer & 0.0897 \\
    2 & Conformer & 0.1786 \\
    3 & ZipFormer & 0.1888 \\
    4 & Multi-ConvFormer & 0.1845 \\
    5 & TASA & \textbf{0.0795} \\ 
    6 & Speech Transformer & 0.2778 \\ \hline
    7 & ViSpeechFormer (ours) & \underline{0.1112} \\ \hline
    \end{tabular}
    \caption{Average speed of of inference (seconds) of baselines and our proposed ViSpeechFormer on the test set of the LSVSC dataset.}
    \label{tab:speed}
\end{table}

We further analyze the computational efficiency of different methods by examining the number of decoder parameters and the average inference speed, as reported in Tables \ref{tab:results} and \ref{tab:speed}. Subword-based baselines generally employ large decoders, with several models exceeding 4M parameters, which leads to higher inference latency. Character-level decoding reduces decoder size, as observed for Speech Transformer, but still results in the slowest inference speed due to longer output sequences. In contrast, ViSpeechFormer employs a phoneme-level decoder with a moderate number of parameters (2.0M), achieving a favorable balance between model capacity and decoding efficiency. 

As shown in Table \ref{tab:speed}, ViSpeechFormer attains an average inference time of 0.1112 seconds, which is faster than most subword-based models and significantly more efficient than character-based decoding, while remaining competitive with the fastest baseline. These results indicate that phoneme-level decoding not only improves recognition performance but also enables efficient inference by reducing vocabulary size and output sequence length, making ViSpeechFormer a practical choice for Vietnamese ASR.

\section{Conclusion}

In this study, we propose ViSpeechFormer, a novel and well linguistic-motivated method for Automatic Speech Recognition in phoneme level for Vietnamese. Experimental results show that ViSpeechFormer is effective compared to character-level and subword-level baselines for ASR on both ViVOS and LSVSC datasets. Moreover, analysis on experimental results shows that ViSpeechFormer has better generalization ability on unseen words and is less biased by the frequency of words in training.

\section*{Acknowledgments}
This research was supported by The VNUHCM-University of Information Technology’s Scientific Research Support Fund.

\section*{Ethical Statement}

There are no ethical issues in our study and experiments.

\bibliographystyle{named}
\bibliography{ijcai26}

@inproceedings{ctc,
author = {Graves, Alex and Fern\'{a}ndez, Santiago and Gomez, Faustino and Schmidhuber, J\"{u}rgen},
title = {Connectionist temporal classification: labelling unsegmented sequence data with recurrent neural networks},
year = {2006},
isbn = {1595933832},
publisher = {Association for Computing Machinery},
address = {New York, NY, USA},
url = {https://doi.org/10.1145/1143844.1143891},
doi = {10.1145/1143844.1143891},
booktitle = {Proceedings of the 23rd International Conference on Machine Learning},
pages = {369–376},
numpages = {8},
location = {Pittsburgh, Pennsylvania, USA},
series = {ICML '06}
}

@article{rnnt,
  title={Sequence Transduction with Recurrent Neural Networks},
  author={Alex Graves},
  journal={ArXiv},
  year={2012},
  volume={abs/1211.3711},
  url={https://api.semanticscholar.org/CorpusID:17194112}
}

@Article{lsvsc,
AUTHOR = {Tran, Linh Thi Thuc and Kim, Han-Gyu and La, Hoang Minh and Van Pham, Su},
TITLE = {Automatic Speech Recognition of Vietnamese for a New Large-Scale Corpus},
JOURNAL = {Electronics},
VOLUME = {13},
YEAR = {2024},
NUMBER = {5},
ARTICLE-NUMBER = {977},
URL = {https://www.mdpi.com/2079-9292/13/5/977},
ISSN = {2079-9292},
DOI = {10.3390/electronics13050977}
}

@inproceedings{conformer,
  title     = {Conformer: Convolution-augmented Transformer for Speech Recognition},
  author    = {Anmol Gulati and James Qin and Chung-Cheng Chiu and Niki Parmar and Yu Zhang and Jiahui Yu and Wei Han and Shibo Wang and Zhengdong Zhang and Yonghui Wu and Ruoming Pang},
  year      = {2020},
  booktitle = {Interspeech 2020},
  pages     = {5036--5040},
  doi       = {10.21437/Interspeech.2020-3015},
  issn      = {2958-1796},
}

@INPROCEEDINGS{speech-transformer,
  author={zhao, Yuanyuan and Li, Jie and Wang, Xiaorui and Li, Yan},
  booktitle={ICASSP 2019 - 2019 IEEE International Conference on Acoustics, Speech and Signal Processing (ICASSP)}, 
  title={The Speechtransformer for Large-scale Mandarin Chinese Speech Recognition}, 
  year={2019},
  volume={},
  number={},
  pages={7095-7099},
  doi={10.1109/ICASSP.2019.8682586}}

@inproceedings{zipformer,
 author = {Yao, Zengwei and Guo, Liyong and Yang, Xiaoyu and Kang, Wei and Kuang, Fangjun and Yang, Yifan and Jin, Zengrui and Lin, Long and Povey, Daniel},
 booktitle = {International Conference on Representation Learning},
 editor = {B. Kim and Y. Yue and S. Chaudhuri and K. Fragkiadaki and M. Khan and Y. Sun},
 pages = {44440--44455},
 title = {Zipformer: A faster and better encoder for automatic speech recognition},
 url = {https://proceedings.iclr.cc/paper_files/paper/2024/file/c1bb0e3b062f0a443f2cc8a4ec4bb30d-Paper-Conference.pdf},
 volume = {2024},
 year = {2024}
}

@article{conv-transformer,
  title={Transformers with convolutional context for ASR},
  author={Mohamed, Abdelrahman and Okhonko, Dmytro and Zettlemoyer, Luke},
  journal={arXiv preprint arXiv:1904.11660},
  year={2019}
}

@inproceedings{multi-convformer,
  title     = {{MULTI-CONVFORMER: Extending Conformer with Multiple Convolution Kernels}},
  author    = {Darshan Prabhu and Yifan Peng and Preethi Jyothi and Shinji Watanabe},
  year      = {2024},
  booktitle = {{Interspeech 2024}},
  pages     = {232--236},
  doi       = {10.21437/Interspeech.2024-2384},
  issn      = {2958-1796},
}

@inproceedings{tasa,
  title     = {{Transmitted and Aggregated Self-Attention for Automatic Speech Recognition}},
  author    = {Tian-Hao Zhang and Xinyuan Qian and Feng Chen and Xu-Cheng Yin},
  year      = {2024},
  booktitle = {{Interspeech 2024}},
  pages     = {227--231},
  doi       = {10.21437/Interspeech.2024-2374},
  issn      = {2958-1796},
}

@inproceedings{attention,
  title={Attention is All you Need},
  author={Ashish Vaswani and Noam Shazeer and Niki Parmar and Jakob Uszkoreit and Llion Jones and Aidan N. Gomez and Lukasz Kaiser and Illia Polosukhin},
  booktitle={Neural Information Processing Systems},
  year={2017},
  url={https://api.semanticscholar.org/CorpusID:13756489}
}

@article{seq2seq,
  title={Sequence to sequence learning with neural networks},
  author={Sutskever, Ilya and Vinyals, Oriol and Le, Quoc V},
  journal={Advances in neural information processing systems},
  volume={27},
  year={2014}
}

@book{hao1998,
  title={Tiếng Việt mấy vấn đề ngữ âm - ngữ pháp - ngữ nghĩa},
  author={Cao Xuân Hạo},
  year={1998},
  publisher={Nhà xuất bản Giáo dục Việt Nam}
}

@book{giap2011,
    author = {Nguyễn Thiện Giáp},
    title = {Vấn đề "từ" trong tiếng Việt},
    publisher = {Nhà xuất bản Giáo dục Việt Nam},
    year = {2011}
}

@book{giap2008,
    author = {Nguyễn Thiện Giáp},
    title = {Từ vựng học tiếng Việt},
    publisher = {Nhà xuất bản Giáo dục Việt Nam},
    year = {2008}
}

@book{thuat2016,
    author = {Đoàn Thiện Thuật},
    title = {Ngữ âm tiếng Việt},
    publisher = {Nhà xuất bản Đại học Quốc gia Hà Nội},
    year = {2016}
}

@inproceedings{vivos,
    title = "A non-expert {K}aldi recipe for {V}ietnamese Speech Recognition System",
    author = "Luong, Hieu-Thi  and
      Vu, Hai-Quan",
    editor = "Murakami, Yohei  and
      Lin, Donghui  and
      Ide, Nancy  and
      Pustejovsky, James",
    booktitle = "Proceedings of the Third International Workshop on Worldwide Language Service Infrastructure and Second Workshop on Open Infrastructures and Analysis Frameworks for Human Language Technologies ({WLSI}/{OIAF}4{HLT}2016)",
    month = dec,
    year = "2016",
    address = "Osaka, Japan",
    publisher = "The COLING 2016 Organizing Committee",
    url = "https://aclanthology.org/W16-5207/",
    pages = "51--55"
}

@article{adam,
  title={Adam: A Method for Stochastic Optimization},
  author={Diederik P. Kingma and Jimmy Ba},
  journal={CoRR},
  year={2014},
  volume={abs/1412.6980},
  url={https://api.semanticscholar.org/CorpusID:6628106}
}

@ARTICLE{hmm,
  author={Rabiner, L.R.},
  journal={Proceedings of the IEEE}, 
  title={A tutorial on hidden Markov models and selected applications in speech recognition}, 
  year={1989},
  volume={77},
  number={2},
  pages={257-286},
  doi={10.1109/5.18626}
}

@book{hmm-gmm,
  title={Spoken Language Processing: A Guide to Theory, Algorithm, and System Development},
  author={Huang, X. and Acero, A. and Hon, H.W.},
  isbn={9780130226167},
  lccn={00050196},
  url={https://books.google.com.vn/books?id=reZQAAAAMAAJ},
  year={2001},
  publisher={Prentice Hall PTR}
}

@inproceedings{rnn-asr,
  title={Speech recognition with deep recurrent neural networks},
  author={Graves, Alex and Mohamed, Abdel-rahman and Hinton, Geoffrey},
  booktitle={2013 IEEE international conference on acoustics, speech and signal processing},
  pages={6645--6649},
  year={2013},
  organization={Ieee}
}

@INPROCEEDINGS{las,
  author={Chan, William and Jaitly, Navdeep and Le, Quoc and Vinyals, Oriol},
  booktitle={2016 IEEE International Conference on Acoustics, Speech and Signal Processing (ICASSP)}, 
  title={Listen, attend and spell: A neural network for large vocabulary conversational speech recognition}, 
  year={2016},
  volume={},
  number={},
  pages={4960-4964},
  doi={10.1109/ICASSP.2016.7472621}
}

@ARTICLE{ctc-attention,
  author={Watanabe, Shinji and Hori, Takaaki and Kim, Suyoun and Hershey, John R. and Hayashi, Tomoki},
  journal={IEEE Journal of Selected Topics in Signal Processing}, 
  title={Hybrid CTC/Attention Architecture for End-to-End Speech Recognition}, 
  year={2017},
  volume={11},
  number={8},
  pages={1240-1253},
  doi={10.1109/JSTSP.2017.2763455}
}

@inproceedings{wav2vec2,
author = {Baevski, Alexei and Zhou, Henry and Mohamed, Abdelrahman and Auli, Michael},
title = {wav2vec 2.0: a framework for self-supervised learning of speech representations},
year = {2020},
isbn = {9781713829546},
publisher = {Curran Associates Inc.},
address = {Red Hook, NY, USA},
booktitle = {Proceedings of the 34th International Conference on Neural Information Processing Systems},
articleno = {1044},
numpages = {12},
location = {Vancouver, BC, Canada},
series = {NIPS '20}
}

@inproceedings{advanced-ctc,
  title     = {Advances in Joint CTC-Attention Based End-to-End Speech Recognition with a Deep CNN Encoder and RNN-LM},
  author    = {Takaaki Hori and Shinji Watanabe and Yu Zhang and William Chan},
  year      = {2017},
  booktitle = {Interspeech 2017},
  pages     = {949--953},
  doi       = {10.21437/Interspeech.2017-1296},
  issn      = {2958-1796},
}

@INPROCEEDINGS{rnn-ctc,
  author={Sak, Haşim and Senior, Andrew and Rao, Kanishka and İrsoy, Ozan and Graves, Alex and Beaufays, Françoise and Schalkwyk, Johan},
  booktitle={2015 IEEE International Conference on Acoustics, Speech and Signal Processing (ICASSP)}, 
  title={Learning acoustic frame labeling for speech recognition with recurrent neural networks}, 
  year={2015},
  volume={},
  number={},
  pages={4280-4284},
  doi={10.1109/ICASSP.2015.7178778}}

@article{layernorm,
  title={Layer normalization},
  author={Ba, Jimmy Lei and Kiros, Jamie Ryan and Hinton, Geoffrey E},
  journal={arXiv preprint arXiv:1607.06450},
  year={2016}
}

\end{document}